\documentclass{article}

\usepackage[final]{DLI_2023}
\usepackage{balance}

\usepackage{graphicx} 
\graphicspath{ {./images/} }

\usepackage[utf8]{inputenc} 
\usepackage[T1]{fontenc}    
\usepackage[hidelinks]{hyperref}     
\usepackage{url}            
\usepackage{booktabs}       
\usepackage{amsfonts}       
\usepackage{nicefrac}       
\usepackage{microtype}      
\usepackage{xcolor}         

\usepackage{longtable}

\title{Benchmarking On-Device Machine Learning on Apple Silicon with MLX}

\author{%
  Oluwaseun A. Ajayi
  \And
  Ogundepo Odunayo 
}

\newcommand{\insertdetailedoperationbenchmarkcuda}{

    \begin{longtable}{lllll}
        \toprule
        & \textbf{Operation}                                           & \textbf{cpu} & \textbf{cuda} & \textbf{cuda/cpu speedup}\\
        \midrule
        & Argmax / dim=64x1024x128 axis=0                  &  15.95 &   0.09 & +17038\% \\
        & Argmax / dim=64x1024x128 axis=1                  &   1.05 &   0.11 &   +862\% \\
        & Argmax / dim=64x1024x128 axis=2                  &   0.59 &   0.09 &   +542\% \\
        & Argmax / dim=64x128x1024 axis=2                  &   0.44 &   0.09 &   +399\% \\
        & BCE / dim=1000000 dim=1000000                   &   0.73 &   0.06 &  +1201\% \\
        & BCE / dim=100000x32 dim=100000x32               &   2.28 &   0.13 &  +1609\% \\
        & BCE / dim=100000x64x2 dim=100000x64x2           &  11.64 &   0.44 &  +2526\% \\
        & BCE / dim=128x100000 dim=128x100000             &  11.77 &   0.44 &  +2561\% \\
        & Concat / dim=1000000x64 dim=1000000x32 axis=1    &  29.50 &   1.77 &  +1570\% \\
        & Concat / dim=1000000x64 dim=1000000x128 axis=1   &  61.04 &   3.43 &  +1681\% \\
        & Concat / dim=1000000x64 dim=1000000x64 axis=0    &  36.29 &   2.30 &  +1474\% \\
        & Concat / dim=64x1000000 dim=64x1000000 axis=0    &  36.27 &   2.28 &  +1487\% \\
        & Conv1d / dim=100x256x3 dim=8x3x3                &   0.13 &   0.05 &   +148\% \\
        & Conv1d / dim=100x256x256 dim=8x3x256            &   4.52 &   0.48 &   +848\% \\
        & Conv1d / dim=16x1000x80 dim=128x11x80           &   1.06 &   0.19 &   +454\% \\
        & Conv1d / dim=16x1000x3 dim=128x11x3             &   0.18 &   0.07 &   +154\% \\
        & Conv2d / dim=100x256x256x3 dim=8x3x3x3          &  16.08 &   2.27 &   +608\% \\
        & Conv2d / dim=10x256x256x12 dim=8x3x3x12         &   0.90 &   0.35 &   +158\% \\
        & Conv2d / dim=1x256x256x128 dim=8x3x3x128        &   0.96 &   0.81 &    +18\% \\
        & Conv2d / dim=100x28x28x3 dim=8x3x3x3            &   0.10 &   0.04 &   +135\% \\
        & Conv2d / dim=1000x28x28x3 dim=8x3x3x3           &   0.39 &   0.22 &    +79\% \\
        & Gather / dim=64x256 dim=10                      &   0.01 &   0.02 &    -50\% \\
        & Gather / dim=64x256 dim=1000                    &   0.02 &   0.02 &    -17\% \\
        & Gather / dim=64x256 dim=1000000                 &  73.86 &   2.52 &  +2832\% \\
        & Gather / dim=1024x32 dim=10                     &   0.01 &   0.02 &    -64\% \\
        & Gather / dim=1024x32 dim=1000                   &   0.01 &   0.02 &    -35\% \\
        & Gather / dim=1024x32 dim=1000000                &   9.44 &   0.35 &  +2603\% \\
        & LeakyReLU / dim=128x16x1024                     &   0.03 &   0.05 &    -34\% \\
        & LeakyReLU / dim=64x128x1024                     &   2.48 &   0.15 &  +1510\% \\
        & Linear / dim=100x1024x32 dim=32x1024 dim=1024   &  37.08 &   1.57 &  +2260\% \\
        & Linear / dim=100x1024x64 dim=64x1024 dim=1024   &  37.77 &   1.87 &  +1918\% \\
        & Linear / dim=100x1024x256 dim=256x1024 dim=1024 &  46.01 &   4.68 &   +883\% \\
        & Linear / dim=100x1024x512 dim=512x1024 dim=1024 &  61.35 &   7.35 &   +734\% \\
        & Linear / dim=100x1x51200 dim=51200x1 dim=1      &   0.04 &   0.08 &    -48\% \\
        & MatMul / dim=32x1x1000 dim=32x1000x128          &   0.04 &   0.06 &    -38\% \\
        & MatMul / dim=1000x64x256 dim=256x32             &   0.54 &   0.20 &   +174\% \\
        & MatMul / dim=1000x64x1024 dim=1000x1024x32      &   2.66 &   2.27 &    +17\% \\
        & MatMul / dim=1000x1024x64 dim=1000x64x256       &  76.78 &   3.76 &  +1939\% \\
        & MatMul / dim=64x1000000 dim=1000000x32          &   2.80 &   0.96 &   +191\% \\
        & MatMul / dim=1000000x64 dim=64x1024             & 297.09 &  16.51 &  +1699\% \\
        & PReLU / dim=128x16x1024 dim=1                   &   0.02 &   0.06 &    -64\% \\
        & PReLU / dim=64x128x1024 dim=1                   &   2.47 &   0.17 &  +1377\% \\
        & ReLU / dim=128x16x1024                          &   0.02 &   0.05 &    -64\% \\
        & ReLU / dim=64x128x1024                          &   2.46 &   0.16 &  +1413\% \\
        & Scatter / dim=64x16 dim=10                      &   0.01 &   0.02 &    -60\% \\
        & Scatter / dim=64x16 dim=1000                    &   0.02 &   0.02 &    -11\% \\
        & Scatter / dim=64x16 dim=1000000                 &   3.94 &   0.20 &  +1845\% \\
        & Scatter / dim=1024x32 dim=10                    &   0.01 &   0.02 &    -60\% \\
        & Scatter / dim=1024x32 dim=1000                  &   0.02 &   0.02 &    -22\% \\
        & Scatter / dim=1024x32 dim=1000000               &   1.85 &   0.35 &   +420\% \\
        & ScatterSum / dim=64x16 dim=10                   &   0.01 &   0.03 &    -75\% \\
        & ScatterSum / dim=64x16 dim=1000                 &   0.01 &   0.03 &    -63\% \\
        & ScatterSum / dim=64x16 dim=1000000              &   5.18 &   0.16 &  +3152\% \\
        & ScatterSum / dim=1024x32 dim=10                 &   0.01 &   0.03 &    -56\% \\
        & ScatterSum / dim=1024x32 dim=1000               &   0.02 &   0.03 &    -44\% \\
        & ScatterSum / dim=1024x32 dim=1000000            &  10.59 &   0.13 &  +8286\% \\
        & ScatterMax / dim=64x16 dim=10                   &   0.01 &   0.03 &    -75\% \\
        & ScatterMax / dim=64x16 dim=1000                 &   0.01 &   0.04 &    -70\% \\
        & ScatterMax / dim=64x16 dim=1000000              &   5.17 &   0.22 &  +2273\% \\
        & ScatterMax / dim=1024x32 dim=10                 &   0.01 &   0.03 &    -58\% \\
        & ScatterMax / dim=1024x32 dim=1000               &   0.02 &   0.03 &    -45\% \\
        & ScatterMax / dim=1024x32 dim=1000000            &  10.36 &   0.14 &  +7079\% \\
        & SeLU / dim=128x16x1024                          &   0.06 &   0.05 &    +20\% \\
        & SeLU / dim=64x128x1024                          &   2.47 &   0.15 &  +1511\% \\
        & Sigmoid / dim=128x16x1024                       &   0.05 &   0.05 &     +0\% \\
        & Sigmoid / dim=64x128x1024                       &   2.48 &   0.15 &  +1519\% \\
        & Softmax / dim=64x1000000 axis=-1                 &  24.21 &   2.12 &  +1041\% \\
        & Softmax / dim=1000000x64 axis=-1                 &  18.72 &   1.08 &  +1630\% \\
        & Softmax / dim=64x16x32x1024 axis=-1              &  10.37 &   0.59 &  +1657\% \\
        & Softmax / dim=128x16x32x1024 axis=-1             &  20.55 &   1.14 &  +1704\% \\
        & Softmax / dim=1024x16x32x128 axis=-1             &  19.62 &   1.14 &  +1625\% \\
        & Softmax / dim=1024x64x32x8 axis=-1               &  14.23 &   0.30 &  +4607\% \\
        & Softplus / dim=128x16x1024                      &   0.14 &   0.05 &   +172\% \\
        & Softplus / dim=64x128x1024                      &   2.55 &   0.15 &  +1559\% \\
        & Sort / dim=64x128x1024 axis=0                    &  22.52 &   3.60 &   +525\% \\
        & Sort / dim=64x128x1024 axis=1                    &  23.00 &   1.84 &  +1152\% \\
        & Sort / dim=64x128x1024 axis=2                    &  22.60 &   0.98 &  +2201\% \\
        & Sum / dim=64x128x128x128 axis=0                  &   3.95 &   1.11 &   +255\% \\
        & Sum / dim=64x128x128x128 axis=1                  &   4.83 &   1.11 &   +333\% \\
        & Sum / dim=64x128x128x128 axis=2                  &   3.29 &   1.09 &   +200\% \\
        & Sum / dim=64x128x128x128 axis=3                  &   3.29 &   1.04 &   +216\% \\
        & SumAll / dim=64x128x128x128                     &   3.00 &   1.08 &   +177\% \\
        & SumAll / dim=1000000                            &   0.03 &   0.03 &     +1\% \\
        & SumAll / dim=1000000x128                        &   2.84 &   1.02 &   +177\% \\
        & SumAll / dim=128x1000000                        &   2.87 &   1.03 &   +179\% \\
        \bottomrule
    \end{longtable}
}

\newcommand{\insertavgoperationbenchmarkcuda}{

    \begin{longtable}{lllll}
        \toprule
            
            & \textbf{Operation}       & \textbf{cpu} & \textbf{cuda} & \textbf{cuda/cpu speedup} \\
            \midrule
            & Argmax      &   4.51 &   0.10 &  +4625\% \\
            & BCE         &   6.60 &   0.27 &  +2358\% \\
            & Concat      &  40.77 &   2.45 &  +1567\% \\
            & Conv1d      &   1.47 &   0.20 &   +644\% \\
            & Conv2d      &   3.69 &   0.74 &   +399\% \\
            & Gather      &  13.89 &   0.49 &  +2717\% \\
            & LeakyReLU   &   1.26 &   0.10 &  +1124\% \\
            & Linear      &  36.45 &   3.11 &  +1072\% \\
            & MatMul      &  63.32 &   3.96 &  +1499\% \\
            & PReLU       &   1.25 &   0.11 &  +1001\% \\
            & ReLU        &   1.24 &   0.11 &  +1052\% \\
            & Scatter     &   0.97 &   0.11 &   +817\% \\
            & ScatterSum  &   2.64 &   0.07 &  +3706\% \\
            & ScatterMax  &   2.60 &   0.08 &  +3026\% \\
            & SeLU        &   1.27 &   0.10 &  +1140\% \\
            & Sigmoid     &   1.26 &   0.10 &  +1143\% \\
            & Softmax     &  17.95 &   1.06 &  +1590\% \\
            & Softplus    &   1.34 &   0.10 &  +1214\% \\
            & Sort        &  22.71 &   2.14 &   +961\% \\
            & Sum         &   3.84 &   1.09 &   +252\% \\
            & SumAll      &   2.18 &   0.79 &   +176\% \\
        \bottomrule
    \end{longtable}
}

\newcommand{\insertdetailedinferencebenchmarkcuda}{

    \begin{longtable}{lllll}
        \toprule
            
            & \textbf{Model}       & \textbf{cpu} & \textbf{cuda} & \textbf{cuda/cpu speedup} \\
            \midrule
            & bert-base-uncased / inputs-char-no=50 / batch-size=1    &  41.82 &  35.37 &    +18\% \\
            & bert-base-uncased / inputs-char-no=50 / batch-size=16   &  51.83 &   8.93 &   +480\% \\
            & bert-base-uncased / inputs-char-no=50 / batch-size=32   &  81.15 &  14.82 &   +447\% \\
            & bert-base-uncased / inputs-char-no=100 / batch-size=1   &  14.26 &   5.39 &   +164\% \\
            & bert-base-uncased / inputs-char-no=100 / batch-size=16  &  59.13 &  10.73 &   +451\% \\
            & bert-base-uncased / inputs-char-no=100 / batch-size=32  & 103.31 &  19.67 &   +425\% \\
            & bert-base-uncased / inputs-char-no=200 / batch-size=1   &  16.46 &   5.40 &   +204\% \\
            & bert-base-uncased / inputs-char-no=200 / batch-size=16  &  93.48 &  17.73 &   +427\% \\
            & bert-base-uncased / inputs-char-no=200 / batch-size=32  & 177.00 &  38.19 &   +363\% \\
            & bert-base-uncased / inputs-char-no=500 / batch-size=1   &  22.38 &   5.45 &   +310\% \\
            & bert-base-uncased / inputs-char-no=500 / batch-size=16  & 183.51 &  40.76 &   +350\% \\
            & bert-base-uncased / inputs-char-no=500 / batch-size=32  & 355.04 &  79.08 &   +348\% \\
            & bert-large-uncased / inputs-char-no=50 / batch-size=1   &  31.41 &   9.99 &   +214\% \\
            & bert-large-uncased / inputs-char-no=50 / batch-size=16  & 176.72 &  24.65 &   +616\% \\
            & bert-large-uncased / inputs-char-no=50 / batch-size=32  & 266.41 &  46.27 &   +475\% \\
            & bert-large-uncased / inputs-char-no=100 / batch-size=1  &  39.19 &  10.19 &   +284\% \\
            & bert-large-uncased / inputs-char-no=100 / batch-size=16 & 194.62 &  31.75 &   +513\% \\
            & bert-large-uncased / inputs-char-no=100 / batch-size=32 & 316.14 &  63.02 &   +401\% \\
            & bert-large-uncased / inputs-char-no=200 / batch-size=1  &  41.95 &  10.36 &   +304\% \\
            & bert-large-uncased / inputs-char-no=200 / batch-size=16 & 282.63 &  54.36 &   +419\% \\
            & bert-large-uncased / inputs-char-no=200 / batch-size=32 & 528.31 & 118.32 &   +346\% \\
            & bert-large-uncased / inputs-char-no=500 / batch-size=1  &  66.32 &  11.07 &   +498\% \\
            & bert-large-uncased / inputs-char-no=500 / batch-size=16 & 553.55 & 127.23 &   +335\% \\
            & bert-large-uncased / inputs-char-no=500 / batch-size=32 & 989.03 & 243.37 &   +306\% \\
            & roberta-base / inputs-char-no=50 / batch-size=1         &  11.84 &   9.63 &    +22\% \\
            & roberta-base / inputs-char-no=50 / batch-size=16        &  44.48 &   7.15 &   +522\% \\
            & roberta-base / inputs-char-no=50 / batch-size=32        &  64.14 &   9.16 &   +600\% \\
            & roberta-base / inputs-char-no=100 / batch-size=1        &  13.10 &   6.86 &    +90\% \\
            & roberta-base / inputs-char-no=100 / batch-size=16       &  76.31 &  10.96 &   +596\% \\
            & roberta-base / inputs-char-no=100 / batch-size=32       & 130.04 &  20.75 &   +526\% \\
            & roberta-base / inputs-char-no=200 / batch-size=1        &  15.36 &   6.84 &   +124\% \\
            & roberta-base / inputs-char-no=200 / batch-size=16       & 181.85 &  32.77 &   +454\% \\
            & roberta-base / inputs-char-no=200 / batch-size=32       & 346.24 &  62.77 &   +451\% \\
            & roberta-base / inputs-char-no=500 / batch-size=1        &  20.52 &   6.98 &   +193\% \\
            & roberta-base / inputs-char-no=500 / batch-size=16       & 962.44 & 104.93 &   +817\% \\
            & roberta-base / inputs-char-no=500 / batch-size=32       & 2171.22 & 190.12 &  +1042\% \\
            & xlm-roberta-base / inputs-char-no=50 / batch-size=1     &  11.62 &   6.95 &    +67\% \\
            & xlm-roberta-base / inputs-char-no=50 / batch-size=16    &  43.34 &   7.27 &   +495\% \\
            & xlm-roberta-base / inputs-char-no=50 / batch-size=32    &  63.46 &   9.09 &   +597\% \\
            & xlm-roberta-base / inputs-char-no=100 / batch-size=1    &  13.43 &   6.92 &    +94\% \\
            & xlm-roberta-base / inputs-char-no=100 / batch-size=16   &  57.67 &   7.36 &   +683\% \\
            & xlm-roberta-base / inputs-char-no=100 / batch-size=32   &  85.66 &  14.37 &   +495\% \\
            & xlm-roberta-base / inputs-char-no=200 / batch-size=1    &  15.63 &   6.93 &   +125\% \\
            & xlm-roberta-base / inputs-char-no=200 / batch-size=16   &  83.88 &  12.24 &   +585\% \\
            & xlm-roberta-base / inputs-char-no=200 / batch-size=32   & 143.51 &  25.78 &   +456\% \\
            & xlm-roberta-base / inputs-char-no=500 / batch-size=1    &  21.53 &   7.01 &   +207\% \\
            & xlm-roberta-base / inputs-char-no=500 / batch-size=16   & 181.72 &  30.58 &   +494\% \\
            & xlm-roberta-base / inputs-char-no=500 / batch-size=32   & 321.61 &  58.55 &   +449\% \\
        \bottomrule
    \end{longtable}
}

\newcommand{\insertavginferenceinpbenchmarkcuda}{

    \begin{longtable}{lllll}
        \toprule
            
        & \textbf{Model}       & \textbf{cpu} & \textbf{cuda} & \textbf{cuda/cpu speedup} \\
        \midrule
        & bert-base-uncased / inputs-char-no=50   &  58.27 &  19.71 &   +195\% \\
        & bert-base-uncased / inputs-char-no=100  &  58.90 &  11.93 &   +393\% \\
        & bert-base-uncased / inputs-char-no=200  &  95.65 &  20.44 &   +367\% \\
        & bert-base-uncased / inputs-char-no=500  & 186.98 &  41.76 &   +347\% \\
        & bert-large-uncased / inputs-char-no=50  & 158.18 &  26.97 &   +486\% \\
        & bert-large-uncased / inputs-char-no=100 & 183.32 &  34.99 &   +423\% \\
        & bert-large-uncased / inputs-char-no=200 & 284.30 &  61.01 &   +365\% \\
        & bert-large-uncased / inputs-char-no=500 & 536.30 & 127.22 &   +321\% \\
        & roberta-base / inputs-char-no=50        &  40.15 &   8.64 &   +364\% \\
        & roberta-base / inputs-char-no=100       &  73.15 &  12.86 &   +469\% \\
        & roberta-base / inputs-char-no=200       & 181.15 &  34.13 &   +430\% \\
        & roberta-base / inputs-char-no=500       & 1051.39 & 100.68 &   +944\% \\
        & xlm-roberta-base / inputs-char-no=50    &  39.47 &   7.77 &   +407\% \\
        & xlm-roberta-base / inputs-char-no=100   &  52.25 &   9.55 &   +447\% \\
        & xlm-roberta-base / inputs-char-no=200   &  81.01 &  14.99 &   +440\% \\
        & xlm-roberta-base / inputs-char-no=500   & 174.95 &  32.05 &   +445\% \\

        \bottomrule
    \end{longtable}
}

\newcommand{\insertavginferencebsizebenchmarkcuda}{

    \begin{longtable}{lllll}
        \toprule
            
        & \textbf{Model}       & \textbf{cpu} & \textbf{cuda} & \textbf{cuda/cpu speedup} \\
        \midrule
        & bert-base-uncased / batch-size=1   &  23.73 &  12.90 &    +83\% \\
        & bert-base-uncased / batch-size=16  &  96.99 &  19.54 &   +396\% \\
        & bert-base-uncased / batch-size=32  & 179.12 &  37.94 &   +372\% \\
        & bert-large-uncased / batch-size=1  &  44.72 &  10.40 &   +329\% \\
        & bert-large-uncased / batch-size=16 & 301.88 &  59.49 &   +407\% \\
        & bert-large-uncased / batch-size=32 & 524.97 & 117.75 &   +345\% \\
        & roberta-base / batch-size=1        &  15.20 &   7.58 &   +100\% \\
        & roberta-base / batch-size=16       & 316.27 &  38.95 &   +711\% \\
        & roberta-base / batch-size=32       & 677.91 &  70.70 &   +858\% \\
        & xlm-roberta-base / batch-size=1    &  15.55 &   6.95 &   +123\% \\
        & xlm-roberta-base / batch-size=16   &  91.65 &  14.36 &   +538\% \\
        & xlm-roberta-base / batch-size=32   & 153.56 &  26.95 &   +469\% \\

        \bottomrule
    \end{longtable}
}

\newcommand{\insertavgmodelbenchmarkcuda}{

    \begin{longtable}{lllll}
        \toprule
            
        & \textbf{Model}       & \textbf{cpu} & \textbf{cuda} & \textbf{cuda/cpu speedup} \\
        \midrule
        & bert-base-uncased  &  99.95 &  23.46 &   +326\% \\
        & bert-large-uncased & 290.52 &  62.55 &   +364\% \\
        & roberta-base       & 336.46 &  39.08 &   +761\% \\
        & xlm-roberta-base   &  86.92 &  16.09 &   +440\% \\
        \bottomrule
    \end{longtable}
}

\newcommand{\insertdetailedoperationbenchmarkmlx}{

    \begin{longtable}{lllll}
        \toprule
        & \textbf{Operation}                                           & \textbf{mlx-gpu} & \textbf{mlx-cpu} & \textbf{mlx-gpu/mlx-cpu speedup}\\
        \midrule
        & Argmax / dim=64x1024x128 axi=0                  &   1.68 &  10.63 &   +531\% \\
        & Argmax / dim=64x1024x128 axi=1                  &   1.77 &  10.60 &   +499\% \\
        & Argmax / dim=64x1024x128 axi=2                  &   1.86 &  10.83 &   +483\% \\
        & Argmax / dim=64x128x1024 axi=2                  &   1.98 &  10.84 &   +446\% \\
        & BCE / dim=1000000 dim=1000000                   &   1.29 &   8.08 &   +526\% \\
        & BCE / dim=100000x32 dim=100000x32               &   3.25 &  26.29 &   +708\% \\
        & BCE / dim=100000x64x2 dim=100000x64x2           &  10.40 & 105.79 &   +917\% \\
        & BCE / dim=128x100000 dim=128x100000             &  10.42 & 105.04 &   +908\% \\
        & Concat / dim=1000000x64 dim=1000000x32 axi=1    &  15.48 &  63.56 &   +310\% \\
        & Concat / dim=1000000x64 dim=1000000x128 axi=1   &  29.11 & 153.33 &   +426\% \\
        & Concat / dim=1000000x64 dim=1000000x64 axi=0    &  17.59 &  61.54 &   +249\% \\
        & Concat / dim=64x1000000 dim=64x1000000 axi=0    &  17.32 &  82.39 &   +375\% \\
        & Conv1d / dim=100x256x3 dim=8x3x3                &   0.51 &   0.36 &    -29\% \\
        & Conv1d / dim=100x256x256 dim=8x3x256            &   6.96 &   9.65 &    +38\% \\
        & Conv1d / dim=16x1000x80 dim=128x11x80           &   5.42 &   6.68 &    +23\% \\
        & Conv1d / dim=16x1000x3 dim=128x11x3             &   0.77 &   0.56 &    -27\% \\
        & Conv2d / dim=100x256x256x3 dim=8x3x3x3          &  17.56 & 1307.08 &  +7345\% \\
        & Conv2d / dim=10x256x256x12 dim=8x3x3x12         &  35.44 & 537.68 &  +1417\% \\
        & Conv2d / dim=1x256x256x128 dim=8x3x3x128        &   1.92 & 734.92 & +38218\% \\
        & Conv2d / dim=100x28x28x3 dim=8x3x3x3            &   0.40 &  14.20 &  +3433\% \\
        & Conv2d / dim=1000x28x28x3 dim=8x3x3x3           &   2.09 & 142.41 &  +6710\% \\
        & Gather / dim=64x256 dim=10                      &   0.18 &   0.02 &    -89\% \\
        & Gather / dim=64x256 dim=1000                    &   0.24 &   0.04 &    -81\% \\
        & Gather / dim=64x256 dim=1000000                 &  32.23 &  24.44 &    -24\% \\
        & Gather / dim=1024x32 dim=10                     &   0.24 &   0.02 &    -91\% \\
        & Gather / dim=1024x32 dim=1000                   &   0.21 &   0.03 &    -87\% \\
        & Gather / dim=1024x32 dim=1000000                &   4.56 &   6.75 &    +48\% \\
        & LeakyReLU / dim=128x16x1024                     &   0.56 &   0.33 &    -40\% \\
        & LeakyReLU / dim=64x128x1024                     &   1.44 &   1.31 &     -8\% \\
        & Linear / dim=100x1024x32 dim=32x1024 dim=1024   &   7.89 &  51.15 &   +547\% \\
        & Linear / dim=100x1024x64 dim=64x1024 dim=1024   &   8.55 &  50.07 &   +485\% \\
        & Linear / dim=100x1024x256 dim=256x1024 dim=1024 &  26.36 &  78.06 &   +196\% \\
        & Linear / dim=100x1024x512 dim=512x1024 dim=1024 &  50.78 & 124.49 &   +145\% \\
        & Linear / dim=100x1x51200 dim=51200x1 dim=1      &   0.83 &   0.56 &    -31\% \\
        & MatMul / dim=32x1x1000 dim=32x1000x128          &   0.58 &   0.87 &    +48\% \\
        & MatMul / dim=1000x64x256 dim=256x32             &   1.87 &   3.19 &    +70\% \\
        & MatMul / dim=1000x64x1024 dim=1000x1024x32      &   7.75 &  18.07 &   +133\% \\
        & MatMul / dim=1000x1024x64 dim=1000x64x256       &  25.64 &  53.14 &   +107\% \\
        & MatMul / dim=64x1000000 dim=1000000x32          &  28.11 &  17.88 &    -36\% \\
        & MatMul / dim=1000000x64 dim=64x1024             &  93.18 & 543.30 &   +483\% \\
        & PReLU / dim=128x16x1024 dim=1                   &   0.49 &   1.40 &   +186\% \\
        & PReLU / dim=64x128x1024 dim=1                   &   1.39 &   5.75 &   +314\% \\
        & ReLU / dim=128x16x1024                          &   0.52 &   0.22 &    -57\% \\
        & ReLU / dim=64x128x1024                          &   1.45 &   1.17 &    -19\% \\
        & Scatter / dim=64x16 dim=10                      &   0.15 &   0.02 &    -88\% \\
        & Scatter / dim=64x16 dim=1000                    &   0.17 &   0.08 &    -54\% \\
        & Scatter / dim=64x16 dim=1000000                 &   1.47 &  65.85 &  +4377\% \\
        & Scatter / dim=1024x32 dim=10                    &   0.19 &   0.02 &    -88\% \\
        & Scatter / dim=1024x32 dim=1000                  &   0.18 &   0.15 &    -18\% \\
        & Scatter / dim=1024x32 dim=1000000               &   2.75 & 120.82 &  +4286\% \\
        & ScatterSum / dim=64x16 dim=10                   &   0.04 &   0.02 &    -61\% \\
        & ScatterSum / dim=64x16 dim=1000                 &   0.04 &   0.02 &    -60\% \\
        & ScatterSum / dim=64x16 dim=1000000              &   0.02 &   0.02 &    -34\% \\
        & ScatterSum / dim=1024x32 dim=10                 &   0.02 &   0.02 &    -34\% \\
        & ScatterSum / dim=1024x32 dim=1000               &   0.02 &   0.01 &    -35\% \\
        & ScatterSum / dim=1024x32 dim=1000000            &   0.03 &   0.01 &    -44\% \\
        & ScatterMax / dim=64x16 dim=10                   &   0.02 &   0.02 &    -34\% \\
        & ScatterMax / dim=64x16 dim=1000                 &   0.03 &   0.01 &    -52\% \\
        & ScatterMax / dim=64x16 dim=1000000              &   0.03 &   0.01 &    -49\% \\
        & ScatterMax / dim=1024x32 dim=10                 &   0.03 &   0.01 &    -48\% \\
        & ScatterMax / dim=1024x32 dim=1000               &   0.03 &   0.02 &    -34\% \\
        & ScatterMax / dim=1024x32 dim=1000000            &   0.02 &   0.01 &    -34\% \\
        & SeLU / dim=128x16x1024                          &   0.51 &   2.02 &   +294\% \\
        & SeLU / dim=64x128x1024                          &   1.36 &   8.06 &   +492\% \\
        & Sigmoid / dim=128x16x1024                       &   0.50 &   1.90 &   +277\% \\
        & Sigmoid / dim=64x128x1024                       &   1.48 &   7.66 &   +417\% \\
        & Softmax / dim=64x1000000 axi=-1                 &  34.17 &  61.08 &    +78\% \\
        & Softmax / dim=1000000x64 axi=-1                 &  33.68 &  62.68 &    +86\% \\
        & Softmax / dim=64x16x32x1024 axi=-1              &  18.11 &  32.96 &    +82\% \\
        & Softmax / dim=128x16x32x1024 axi=-1             &  35.76 &  65.80 &    +83\% \\
        & Softmax / dim=1024x16x32x128 axi=-1             &  36.10 &  65.61 &    +81\% \\
        & Softmax / dim=1024x64x32x8 axi=-1               &   9.62 &  16.33 &    +69\% \\
        & Softplus / dim=128x16x1024                      &   0.58 &  14.73 &  +2457\% \\
        & Softplus / dim=64x128x1024                      &   1.56 &  58.23 &  +3640\% \\
        & Sort / dim=64x128x1024 axi=0                    &   3.39 & 271.11 &  +7905\% \\
        & Sort / dim=64x128x1024 axi=1                    &   3.38 & 264.27 &  +7729\% \\
        & Sort / dim=64x128x1024 axi=2                    &   3.37 & 261.29 &  +7661\% \\
        & Sum / dim=64x128x128x128 axi=0                  &   9.00 &   9.20 &     +2\% \\
        & Sum / dim=64x128x128x128 axi=1                  &   8.76 &   9.36 &     +6\% \\
        & Sum / dim=64x128x128x128 axi=2                  &   8.87 &  10.41 &    +17\% \\
        & Sum / dim=64x128x128x128 axi=3                  &   8.71 &   9.22 &     +5\% \\
        & SumAll / dim=64x128x128x128                     &   8.83 &   8.75 &      0\% \\
        & SumAll / dim=1000000                            &   0.30 &   0.07 &    -77\% \\
        & SumAll / dim=1000000x128                        &   8.55 &   8.33 &     -2\% \\
        & SumAll / dim=128x1000000                        &   8.42 &   8.35 &      0\% \\
        \bottomrule
    \end{longtable}
}

\newcommand{\insertavgoperationbenchmarkmlx}{

    \begin{longtable}{lllll}
            \toprule
            
            & \textbf{Operation}       & \textbf{mlx-gpu} & \textbf{mlx-cpu} & \textbf{mlx-gpu/mlx-cpu speedup} \\
            \midrule
            & Argmax      &   1.82 &  10.73 &   +488\% \\
            & BCE         &   6.34 &  61.30 &   +867\% \\
            & Concat      &  19.88 &  90.20 &   +353\% \\
            & Conv1d      &   3.41 &   4.31 &    +26\% \\
            & Conv2d      &  11.48 & 547.26 &  +4666\% \\
            & Gather      &   6.28 &   5.22 &    -16\% \\
            & LeakyReLU   &   1.00 &   0.82 &    -17\% \\
            & Linear      &  18.88 &  60.87 &   +222\% \\
            & MatMul      &  26.19 & 106.08 &   +305\% \\
            & PReLU       &   0.94 &   3.57 &   +281\% \\
            & ReLU        &   0.98 &   0.69 &    -29\% \\
            & Scatter     &   0.82 &  31.16 &  +3697\% \\
            & ScatterSum  &   0.03 &   0.02 &    -48\% \\
            & ScatterMax  &   0.03 &   0.02 &    -43\% \\
            & SeLU        &   0.94 &   5.04 &   +438\% \\
            & Sigmoid     &   0.99 &   4.78 &   +381\% \\
            & Softmax     &  27.91 &  50.74 &    +81\% \\
            & Softplus    &   1.07 &  36.48 &  +3321\% \\
            & Sort        &   3.38 & 265.56 &  +7765\% \\
            & Sum         &   8.83 &   9.55 &     +8\% \\
            & SumAll      &   6.53 &   6.38 &     -2\% \\
        \bottomrule
    \end{longtable}
}

\newcommand{\insertdetailedinferencebenchmarkmlx}{

    \begin{longtable}{lllll}
        \toprule
            
        & \textbf{Model}                                                   & \textbf{mlx-gpu} & \textbf{mlx-cpu} & \textbf{mlx-gpu/mlx-cpu speedup} \\
        \midrule
        & bert-base-uncased / inputs-char-no=50 / batch-size=1    &   14.48 &   41.67 &                 +187.8\% \\
        & bert-base-uncased / inputs-char-no=50 / batch-size=16   &   73.15 &  216.89 &                 +196.5\% \\
        & bert-base-uncased / inputs-char-no=50 / batch-size=32   &  141.51 &  422.21 &                 +198.4\% \\
        & bert-base-uncased / inputs-char-no=100 / batch-size=1   &   14.36 &   38.44 &                 +167.7\% \\
        & bert-base-uncased / inputs-char-no=100 / batch-size=16  &   94.06 &  264.22 &                 +180.9\% \\
        & bert-base-uncased / inputs-char-no=100 / batch-size=32  &  187.27 &  554.67 &                 +196.2\% \\
        & bert-base-uncased / inputs-char-no=200 / batch-size=1   &   16.89 &   46.64 &                 +176.1\% \\
        & bert-base-uncased / inputs-char-no=200 / batch-size=16  &  169.06 &  494.90 &                 +192.7\% \\
        & bert-base-uncased / inputs-char-no=200 / batch-size=32  &  335.41 & 1065.57 &                 +217.7\% \\
        & bert-base-uncased / inputs-char-no=500 / batch-size=1   &   29.91 &   82.48 &                 +175.8\% \\
        & bert-base-uncased / inputs-char-no=500 / batch-size=16  &  365.54 & 1170.29 &                 +220.2\% \\
        & bert-base-uncased / inputs-char-no=500 / batch-size=32  &  710.51 & 2287.70 &                 +222.0\% \\
        & bert-large-uncased / inputs-char-no=50 / batch-size=1   &   42.18 &  124.95 &                 +196.2\% \\
        & bert-large-uncased / inputs-char-no=50 / batch-size=16  &  206.24 &  637.74 &                 +209.2\% \\
        & bert-large-uncased / inputs-char-no=50 / batch-size=32  &  435.64 & 1269.85 &                 +191.5\% \\
        & bert-large-uncased / inputs-char-no=100 / batch-size=1  &   39.66 &  116.85 &                 +194.6\% \\
        & bert-large-uncased / inputs-char-no=100 / batch-size=16 &  267.66 &  768.38 &                 +187.1\% \\
        & bert-large-uncased / inputs-char-no=100 / batch-size=32 &  506.84 & 1611.00 &                 +217.9\% \\
        & bert-large-uncased / inputs-char-no=200 / batch-size=1  &   49.67 &  142.51 &                 +186.9\% \\
        & bert-large-uncased / inputs-char-no=200 / batch-size=16 &  459.70 & 1413.81 &                 +207.6\% \\
        & bert-large-uncased / inputs-char-no=200 / batch-size=32 &  944.39 & 2923.94 &                 +209.6\% \\
        & bert-large-uncased / inputs-char-no=500 / batch-size=1  &  101.30 &  241.95 &                 +138.8\% \\
        & bert-large-uncased / inputs-char-no=500 / batch-size=16 & 1049.63 & 3190.13 &                 +203.9\% \\
        & bert-large-uncased / inputs-char-no=500 / batch-size=32 & 2272.34 & 6262.58 &                 +175.6\% \\
        & roberta-base / inputs-char-no=50 / batch-size=1         &   12.79 &   36.76 &                 +187.4\% \\
        & roberta-base / inputs-char-no=50 / batch-size=16        &   56.81 &  112.73 &                  +98.4\% \\
        & roberta-base / inputs-char-no=50 / batch-size=32        &   96.12 &  212.97 &                 +121.6\% \\
        & roberta-base / inputs-char-no=100 / batch-size=1        &   13.54 &   31.70 &                 +134.1\% \\
        & roberta-base / inputs-char-no=100 / batch-size=16       &  116.20 &  248.24 &                 +113.6\% \\
        & roberta-base / inputs-char-no=100 / batch-size=32       &  211.54 &  511.98 &                 +142.0\% \\
        & roberta-base / inputs-char-no=200 / batch-size=1        &   15.32 &   32.06 &                 +109.3\% \\
        & roberta-base / inputs-char-no=200 / batch-size=16       &  319.59 &  773.45 &                 +142.0\% \\
        & roberta-base / inputs-char-no=200 / batch-size=32       &  629.75 & 1548.77 &                 +145.9\% \\
        & roberta-base / inputs-char-no=500 / batch-size=1        &   25.35 &   50.56 &                  +99.4\% \\
        & roberta-base / inputs-char-no=500 / batch-size=16       & 1108.91 & 2680.44 &                 +141.7\% \\
        & roberta-base / inputs-char-no=500 / batch-size=32       & 2209.37 & 5353.93 &                 +142.3\% \\
        & xlm-roberta-base / inputs-char-no=50 / batch-size=1     &   11.67 &   34.47 &                 +195.4\% \\
        & xlm-roberta-base / inputs-char-no=50 / batch-size=16    &   55.08 &  113.81 &                 +106.6\% \\
        & xlm-roberta-base / inputs-char-no=50 / batch-size=32    &   96.35 &  211.66 &                 +119.7\% \\
        & xlm-roberta-base / inputs-char-no=100 / batch-size=1    &   13.87 &   30.05 &                 +116.7\% \\
        & xlm-roberta-base / inputs-char-no=100 / batch-size=16   &   75.51 &  160.46 &                 +112.5\% \\
        & xlm-roberta-base / inputs-char-no=100 / batch-size=32   &  137.87 &  316.83 &                 +129.8\% \\
        & xlm-roberta-base / inputs-char-no=200 / batch-size=1    &   16.07 &   32.43 &                 +101.8\% \\
        & xlm-roberta-base / inputs-char-no=200 / batch-size=16   &  134.81 &  300.25 &                 +122.7\% \\
        & xlm-roberta-base / inputs-char-no=200 / batch-size=32   &  253.59 &  624.37 &                 +146.2\% \\
        & xlm-roberta-base / inputs-char-no=500 / batch-size=1    &   26.21 &   52.44 &                 +100.1\% \\
        & xlm-roberta-base / inputs-char-no=500 / batch-size=16   &  299.24 &  712.59 &                 +138.1\% \\
        & xlm-roberta-base / inputs-char-no=500 / batch-size=32   &  588.79 & 1412.94 &                 +140.0\% \\
        \bottomrule
    \end{longtable}
}

\newcommand{\insertavginferenceinpbenchmarkmlx}{

    \begin{longtable}{lllll}
        \toprule
            
            & \textbf{Model}                                   & \textbf{mlx-gpu} & \textbf{mlx-cpu} & \textbf{mlx-gpu/mlx-cpu speedup} \\
            \midrule
            & bert-base-uncased / inputs-char-no=50   &   76.38 &  226.92 &                 +197.1\% \\
            & bert-base-uncased / inputs-char-no=100  &   98.56 &  285.78 &                 +189.9\% \\
            & bert-base-uncased / inputs-char-no=200  &  173.78 &  535.71 &                 +208.3\% \\
            & bert-base-uncased / inputs-char-no=500  &  368.65 & 1180.16 &                 +220.1\% \\
            & bert-large-uncased / inputs-char-no=50  &  228.02 &  677.51 &                 +197.1\% \\
            & bert-large-uncased / inputs-char-no=100 &  271.39 &  832.08 &                 +206.6\% \\
            & bert-large-uncased / inputs-char-no=200 &  484.59 & 1493.42 &                 +208.2\% \\
            & bert-large-uncased / inputs-char-no=500 & 1141.09 & 3231.55 &                 +183.2\% \\
            & roberta-base / inputs-char-no=50        &   55.24 &  120.82 &                 +118.7\% \\
            & roberta-base / inputs-char-no=100       &  113.76 &  263.97 &                 +132.0\% \\
            & roberta-base / inputs-char-no=200       &  321.55 &  784.76 &                 +144.1\% \\
            & roberta-base / inputs-char-no=500       & 1114.54 & 2694.98 &                 +141.8\% \\
            & xlm-roberta-base / inputs-char-no=50    &   54.37 &  119.98 &                 +120.7\% \\
            & xlm-roberta-base / inputs-char-no=100   &   75.75 &  169.11 &                 +123.2\% \\
            & xlm-roberta-base / inputs-char-no=200   &  134.83 &  319.02 &                 +136.6\% \\
            & xlm-roberta-base / inputs-char-no=500   &  304.75 &  725.99 &                 +138.2\% \\

        \bottomrule
    \end{longtable}
}

\newcommand{\insertavginferencebsizebenchmarkmlx}{

    \begin{longtable}{lllll}
        \toprule
            
            & \textbf{Model}                              & \textbf{mlx-gpu} & \textbf{mlx-cpu} & \textbf{mlx-gpu/mlx-cpu speedup} \\
            \midrule
            & bert-base-uncased / batch-size=1   &   18.91 &   52.31 &                 +176.6\% \\
            & bert-base-uncased / batch-size=16  &  175.45 &  536.57 &                 +205.8\% \\
            & bert-base-uncased / batch-size=32  &  343.67 & 1082.54 &                 +214.9\% \\
            & bert-large-uncased / batch-size=1  &   58.21 &  156.57 &                 +169.0\% \\
            & bert-large-uncased / batch-size=16 &  495.81 & 1502.52 &                 +203.0\% \\
            & bert-large-uncased / batch-size=32 & 1039.80 & 3016.84 &                 +190.1\% \\
            & roberta-base / batch-size=1        &   16.75 &   37.77 &                 +125.5\% \\
            & roberta-base / batch-size=16       &  400.38 &  953.72 &                 +138.2\% \\
            & roberta-base / batch-size=32       &  786.70 & 1906.91 &                 +142.4\% \\
            & xlm-roberta-base / batch-size=1    &   16.95 &   37.35 &                 +120.4\% \\
            & xlm-roberta-base / batch-size=16   &  141.16 &  321.78 &                 +128.0\% \\
            & xlm-roberta-base / batch-size=32   &  269.15 &  641.45 &                 +138.3\% \\

        \bottomrule
    \end{longtable}
}

\newcommand{\insertavgmodelbenchmarkmlx}{

    \begin{longtable}{lllll}
        \toprule
            
            & \textbf{Model}              & \textbf{mlx-gpu} & \textbf{mlx-cpu} & \textbf{mlx-gpu/mlx-cpu speedup} \\
            \midrule
            & bert-base-uncased  &  179.35 &  557.14 &                 +210.6\% \\
            & bert-large-uncased &  531.27 & 1558.64 &                 +193.4\% \\
            & roberta-base       &  401.27 &  966.13 &                 +140.8\% \\
            & xlm-roberta-base   &  142.42 &  333.52 &                 +134.2\% \\
        \bottomrule
    \end{longtable}
}
            
\newcommand{\insertdetailedinferencebenchmarkmlxmax}{

    \begin{longtable}{lllll}
        \toprule
            
            & \textbf{Model}                                                   & \textbf{mlx-gpu} \\
            \midrule
            & bert-base-uncased / inputs-char-no=50 / batch-size=1    &   4.92 \\
            & bert-base-uncased / inputs-char-no=50 / batch-size=16   &  15.81 \\
            & bert-base-uncased / inputs-char-no=50 / batch-size=32   &  30.06 \\
            & bert-base-uncased / inputs-char-no=100 / batch-size=1   &   9.01 \\
            & bert-base-uncased / inputs-char-no=100 / batch-size=16  &  21.05 \\
            & bert-base-uncased / inputs-char-no=100 / batch-size=32  &  37.77 \\
            & bert-base-uncased / inputs-char-no=200 / batch-size=1   &   8.63 \\
            & bert-base-uncased / inputs-char-no=200 / batch-size=16  &  33.47 \\
            & bert-base-uncased / inputs-char-no=200 / batch-size=32  &  68.89 \\
            & bert-base-uncased / inputs-char-no=500 / batch-size=1   &   9.53 \\
            & bert-base-uncased / inputs-char-no=500 / batch-size=16  &  74.47 \\
            & bert-base-uncased / inputs-char-no=500 / batch-size=32  & 145.19 \\
            & bert-large-uncased / inputs-char-no=50 / batch-size=1   &  10.07 \\
            & bert-large-uncased / inputs-char-no=50 / batch-size=16  &  40.89 \\
            & bert-large-uncased / inputs-char-no=50 / batch-size=32  &  73.22 \\
            & bert-large-uncased / inputs-char-no=100 / batch-size=1  &  16.63 \\
            & bert-large-uncased / inputs-char-no=100 / batch-size=16 &  52.46 \\
            & bert-large-uncased / inputs-char-no=100 / batch-size=32 &  97.90 \\
            & bert-large-uncased / inputs-char-no=200 / batch-size=1  &  15.41 \\
            & bert-large-uncased / inputs-char-no=200 / batch-size=16 &  81.10 \\
            & bert-large-uncased / inputs-char-no=200 / batch-size=32 & 175.80 \\
            & bert-large-uncased / inputs-char-no=500 / batch-size=1  &  20.11 \\
            & bert-large-uncased / inputs-char-no=500 / batch-size=16 & 191.46 \\
            & bert-large-uncased / inputs-char-no=500 / batch-size=32 & 353.62 \\
            & roberta-base / inputs-char-no=50 / batch-size=1         &   3.24 \\
            & roberta-base / inputs-char-no=50 / batch-size=16        &  11.94 \\
            & roberta-base / inputs-char-no=50 / batch-size=32        &  19.78 \\
            & roberta-base / inputs-char-no=100 / batch-size=1        &   6.41 \\
            & roberta-base / inputs-char-no=100 / batch-size=16       &  23.06 \\
            & roberta-base / inputs-char-no=100 / batch-size=32       &  41.55 \\
            & roberta-base / inputs-char-no=200 / batch-size=1        &   5.87 \\
            & roberta-base / inputs-char-no=200 / batch-size=16       &  59.53 \\
            & roberta-base / inputs-char-no=200 / batch-size=32       & 116.36 \\
            & roberta-base / inputs-char-no=500 / batch-size=1        &   6.72 \\
            & roberta-base / inputs-char-no=500 / batch-size=16       & 196.30 \\
            & roberta-base / inputs-char-no=500 / batch-size=32       & 401.05 \\
            & xlm-roberta-base / inputs-char-no=50 / batch-size=1     &   3.32 \\
            & xlm-roberta-base / inputs-char-no=50 / batch-size=16    &  11.49 \\
            & xlm-roberta-base / inputs-char-no=50 / batch-size=32    &  19.29 \\
            & xlm-roberta-base / inputs-char-no=100 / batch-size=1    &   6.60 \\
            & xlm-roberta-base / inputs-char-no=100 / batch-size=16   &  15.67 \\
            & xlm-roberta-base / inputs-char-no=100 / batch-size=32   &  26.45 \\
            & xlm-roberta-base / inputs-char-no=200 / batch-size=1    &   5.86 \\
            & xlm-roberta-base / inputs-char-no=200 / batch-size=16   &  26.06 \\
            & xlm-roberta-base / inputs-char-no=200 / batch-size=32   &  47.55 \\
            & xlm-roberta-base / inputs-char-no=500 / batch-size=1    &   7.11 \\
            & xlm-roberta-base / inputs-char-no=500 / batch-size=16   &  54.37 \\
            & xlm-roberta-base / inputs-char-no=500 / batch-size=32   & 104.63 \\
        \bottomrule
    \end{longtable}
}

\newcommand{\insertavginferenceinpbenchmarkmlxmax}{

    \begin{longtable}{lllll}
        \toprule
            
            & \textbf{Model}                                   & \textbf{mlx-gpu} \\
            \midrule
            & bert-base-uncased / inputs-char-no=50   &  16.93 \\
            & bert-base-uncased / inputs-char-no=100  &  22.61 \\
            & bert-base-uncased / inputs-char-no=200  &  37.00 \\
            & bert-base-uncased / inputs-char-no=500  &  76.40 \\
            & bert-large-uncased / inputs-char-no=50  &  41.40 \\
            & bert-large-uncased / inputs-char-no=100 &  55.66 \\
            & bert-large-uncased / inputs-char-no=200 &  90.77 \\
            & bert-large-uncased / inputs-char-no=500 & 188.39 \\
            & roberta-base / inputs-char-no=50        &  11.65 \\
            & roberta-base / inputs-char-no=100       &  23.67 \\
            & roberta-base / inputs-char-no=200       &  60.59 \\
            & roberta-base / inputs-char-no=500       & 201.36 \\
            & xlm-roberta-base / inputs-char-no=50    &  11.37 \\
            & xlm-roberta-base / inputs-char-no=100   &  16.24 \\
            & xlm-roberta-base / inputs-char-no=200   &  26.49 \\
            & xlm-roberta-base / inputs-char-no=500   &  55.37 \\

        \bottomrule
    \end{longtable}
}

\newcommand{\insertavginferencebsizebenchmarkmlxmax}{

    \begin{longtable}{lllll}
        \toprule
            
            & \textbf{Model}                              & \textbf{mlx-gpu} \\
            \midrule
            & bert-base-uncased / batch-size=1   &   8.02 \\
            & bert-base-uncased / batch-size=16  &  36.20 \\
            & bert-base-uncased / batch-size=32  &  70.48 \\
            & bert-large-uncased / batch-size=1  &  15.56 \\
            & bert-large-uncased / batch-size=16 &  91.48 \\
            & bert-large-uncased / batch-size=32 & 175.13 \\
            & roberta-base / batch-size=1        &   5.56 \\
            & roberta-base / batch-size=16       &  72.71 \\
            & roberta-base / batch-size=32       & 144.68 \\
            & xlm-roberta-base / batch-size=1    &   5.73 \\
            & xlm-roberta-base / batch-size=16   &  26.90 \\
            & xlm-roberta-base / batch-size=32   &  49.48 \\

        \bottomrule
    \end{longtable}
}

\newcommand{\insertavgmodelbenchmarkmlxmax}{

            \begin{longtable}{lllll}
                \toprule
                    
                    & \textbf{Model}              & \textbf{mlx-gpu} \\
                    \midrule
                    & bert-base-uncased  &  38.23 \\
                    & bert-large-uncased &  94.06 \\
                    & roberta-base       &  74.32 \\
                    & xlm-roberta-base   &  27.37 \\
                \bottomrule
            \end{longtable}
    
}
            
\begin{document}

\maketitle

\begin{abstract}




The recent widespread adoption of Large Language Models (LLMs) and machine learning in general has sparked research interest in exploring the possibilities of deploying these models on smaller devices such as laptops and mobile phones.
This creates a need for frameworks and approaches that are capable of taking advantage of on-device hardware.
The MLX framework was created to address this need.
It is a framework optimized for machine learning (ML) computations on Apple silicon devices, facilitating easier research, experimentation, and prototyping.

This paper presents a performance evaluation of MLX, focusing on inference latency of transformer models. We compare the performance of different transformer architecture implementations in MLX with their Pytorch counterparts.
For this research we create a framework called MLX-transformers which includes different transformer implementations in MLX and downloads the model checkpoints in pytorch and converts it to the MLX format. By leveraging the advanced architecture and capabilities of Apple Silicon, MLX-Transformers enables seamless execution of transformer models directly sourced from Hugging Face, eliminating the need for checkpoint conversion often required when porting models between frameworks.

Our study benchmarks different transformer models on two Apple Silicon macbook devices against an NVIDIA CUDA GPU. Specifically, we compare the inference latency performance of models with the same parameter sizes and checkpoints.
We evaluate the performance of BERT, RoBERTa, and XLM-RoBERTa models, with the intention of extending future work to include models of different modalities, thus providing a more comprehensive assessment of MLX's capabilities.
The results highlight MLX's potential in enabling efficient and more accessible on-device ML applications within Apple's ecosystem.

\end{abstract}

\section{INTRODUCTION}
 
The advent of advanced machine learning techniques has increased the utility and applications of language models across various fields. This increase in utility has necessitated the development of more efficient methods to run these models on laptops and other edge devices. 
To address this challenge, researchers have explored various techniques, such as quantization, to enable large models with billions of parameters to run on consumer hardware \citep{Egashira2024}.

Apple's introduction of her proprietary silicon architecture has significantly enhanced the on-device hardware capabilities of its devices\footnote{https://www.apple.com/newsroom/2020/11/introducing-the-next-generation-of-mac/}. 
To fully leverage these capabilities, Apple's machine learning research team introduced the MLX framework\citep{mlx2023}, specifically designed to exploit the GPU capabilities of this hardware for running machine learning models.

MLX \citep{mlx2023} is tailored to the unique architecture of Apple's processors, enhancing both the training and inference phases of machine learning models. 
While inspired by frameworks like PyTorch, Jax, and ArrayFire\footnote{\url{https://ml-explore.github.io/mlx/build/html/index.html}}, MLX differs notably in that it was built specifically to take advantage of a unified memory model. 
The Unified Memory Model (UMM) in Apple's M-series chips fundamentally differs from traditional memory architectures. 
In UMM, the CPU and GPU share a single pool of high-bandwidth memory, allowing both to access the entire memory space directly. 
This contrasts with conventional systems where the CPU uses separate DRAM (main memory) and the GPU uses VRAM, necessitating data transfers between them. 
While both architectures utilize SRAM for fast, local cache access, UMM's shared memory approach is designed to reduces latency and memory duplication. 

The Transformer is a widely used architecture in deep learning, especially for natural language processing (NLP). It uses attention mechanism to process input data in parallel, allowing it to consider the context of each word in a sentence by examining relationships between all words \citep{Vaswani2017}. This architecture has been the foundation for several state-of-the-art models in NLP, including BERT \citep{Devlin2018}, RoBERTa \citep{Liu2019}, XLM-RoBERTa \citep{Conneau2020}, and LLaMA \citep{Touvron2023}. The parallel processing capabilities of GPUs are needed to run transformers with low latency.

This paper introduces MLX-Transformers\footnote{\url{https://github.com/ToluClassics/mlx-transformers}}, a library that integrates transformer models within the MLX framework, specifically designed for Apple Silicon devices. MLX-Transformer simplifies the deployment of transformers by allowing models to run natively on Apple hardware without the need for format conversions or additional compatibility layers. It draws inspiration from the Transformers library \citep{Wolf2020} but is tailored for MLX support.

One of the key advantages of MLX-Transformers is its ability to download and execute transformer models directly from popular platforms like Hugging Face Hub, significantly reducing barriers to entry for developers and researchers. MLX-Transformers currently supports various transformer models, including BERT, Fuyu, LLaMA, M2M-100, OpenELM, Persimmon, Phi, Phi 3, RoBERTa, T5, and XLM-RoBERTa.

In this paper, we present an evaluation and benchmarking of MLX-Transformers' performance on three of its supported models: BERT, RoBERTa, and XLM-RoBERTa. We compare its computational efficiency across various Apple Silicon devices, including the latest M1 and M2, against a CUDA-enabled GPU - NVIDIA A10. 

We present the benchmark results that highlight the performance of MLX-Transformers across different hardware configurations and we hope that the findings from this paper would be useful to the advancement of machine learning on consumer devices.


\section{Related Work}

\textbf{Apple Silicon Performance in Scientific
Computing}

\citep{kenyon2022applesiliconperformancescientific} explores the potential of Apple Silicon processors, particularly the M1 and M1 Ultra, in scientific computing. It compares their performance to state-of-the-art data-center GPUs, such as the NVIDIA V100 and A100, using the SHOC benchmark suite with OpenCL benchmarks. The study finds that Apple Silicon processors outperform these GPUs in single-precision computing tasks, despite lacking double precision GPU capabilities, and emphasize its cost-effectiveness and promising performance in scientific research applications.

This work is relevant to our research as it underscores the efficacy of Apple Silicon in computationally intensive tasks, aligning with our focus on benchmarking the MLX's performance on Apple Silicon for machine learning workloads. While their study is centered on scientific computing benchmarks, our work extends the investigation to machine learning applications, particularly in the context of transformer models.

\textbf{MLX-Based Libraries and Frameworks}

The release of MLX by the Apple Machine Learning Research Team in late 2023 has sparked a surge of research and development efforts aimed at leveraging the framework's capabilities on Apple Silicon devices. Several notable libraries and frameworks have emerged in the wake of MLX's introduction, each contributing to the exploration of MLX's efficiency and potential. 
\textbf{\href{https://github.com/argmaxinc/DiffusionKit}{DiffusionKit}}, for instance, is a library that facilitates the running of Diffusion Models on Apple Silicon with Core ML and MLX. 
It comprises two packages: the first, \textit{diffusionkit}, is a Python package that enables the conversion of PyTorch models to Core ML format, allowing for image generation using MLX in Python, while the second, \textit{DiffusionKit}, is a Swift package designed for on-device inference of diffusion models using Core ML and MLX. 
Another example is \textbf{\href{https://github.com/filipstrand/mflux}{MFLUX}} (MacFLUX), a project that represents a line-by-line port of the FLUX implementation from the Huggingface Diffusers library\footnote{\url{https://github.com/huggingface/diffusers}} to Apple MLX \citep{mflux}, showcasing efforts to adapt established diffusion model implementations to the MLX framework for potential performance benefits on Apple Silicon. 
\textbf{\href{https://github.com/Blaizzy/mlx-vlm}{MLX-VLM}} is another MLX-Based package that enables the execution of Vision Language Models (VLMs) on Apple Silicon devices using MLX \citep{mlx-vlm}. 
\textbf{\href{https://github.com/Blaizzy/fastmlx}{FastMLX}}, a production-ready API that facilitates the hosting of MLX models, including both Vision Language Models (VLMs) and Language Models (LMs) \citep{fastmlx}. 
Furthermore, \href{https://localchat.co/}{\textbf{Local Chat}}, a private AI chatbot built on MLX, allowing users to run the latest open language models on Mac, iPad, and iPhone.

These diverse projects collectively illustrate the growing ecosystem around MLX, spanning image generation, vision-language models and educational resources. They underscore the community's efforts to explore and expand the capabilities of machine learning on Apple Silicon devices using the MLX. As research in this area continues to evolve, these projects lay the groundwork for further innovations and optimizations in on-device machine learning.

\section{Experiments and Experiment Setup}

Our experiments aimed to benchmark the latency performance of MLX operations and transformer models on various hardware configurations. We conducted two main types of benchmarks: operations benchmarking and model inference benchmarking. Both experiments were performed on three devices: (1) 8GB Apple M1 Macbook pro, (2) 32GB Apple M2 Max Macbook pro, and (3) a CUDA-enabled GPU - NVIDIA A10 (24 GB PCIe) on an AWS EC2 instance with 30 vCPUs, 205.4 GB RAM, and 1.5 TB SSD.

For both benchmarks, we conducted multiple iterations to ensure reliability and minimize variability. We presented our results in two formats: Detailed Benchmarks, which provide runtime for each individual experiment, and Average Runtime Benchmarks, which calculate the mean runtime across iterations. The comprehensive results are available in the appendices.

\subsection{Benchmarking Operations}

For operations benchmarking, we utilized code from the \href{https://github.com/TristanBilot/mlx-benchmark}{mlx-benchmark} GitHub Repository \citep{Bilot2024}. We evaluated each MLX operation on both GPU and CPU of Apple Silicon devices and compared these results with their PyTorch equivalents running on CPU and CUDA GPUs. Each test was conducted over five iterations.

The Detailed Benchmark results for operations are presented in Appendix A, while the Average Runtime Benchmark results are in Appendix B. These results offer insights into the performance of individual operations across different hardware configurations.

\subsection{Benchmarking Model Inference}

For model inference benchmarking, we focused on three transformer models supported by MLX-Transformers: BERT, RoBERTa, and XLM-RoBERTa. We measured inference times across different backends (mlx-cpu and mlx-gpu on Apple Silicon; torch-cpu and torch-cuda on CUDA-Enabled GPU) over ten iterations.

To simulate diverse usage scenarios, we generated inputs of various character lengths (50, 100, 200, and 500) and batch sizes (1, 16, and 32) using datasets from the Hugging Face library. This approach allowed us to analyze the influence of input length and batch size on inference performance.

The detailed runtime benchmarks for model inference are presented in Appendix C, while Appendices D, E and F summarize the average runtime benchmarks across different input lengths and batch sizes. This structured approach captures both detailed and high-level performance characteristics of the models under various conditions.

These experiments provide comprehensive insights into the performance dynamics of MLX operations and transformer models across different hardware configurations, input sizes, and batch sizes.

\section{Results and Discussion}

\subsection{Benchmarking Operations}
\begin{figure}[h]
\centering
\includegraphics[width=1\textwidth]{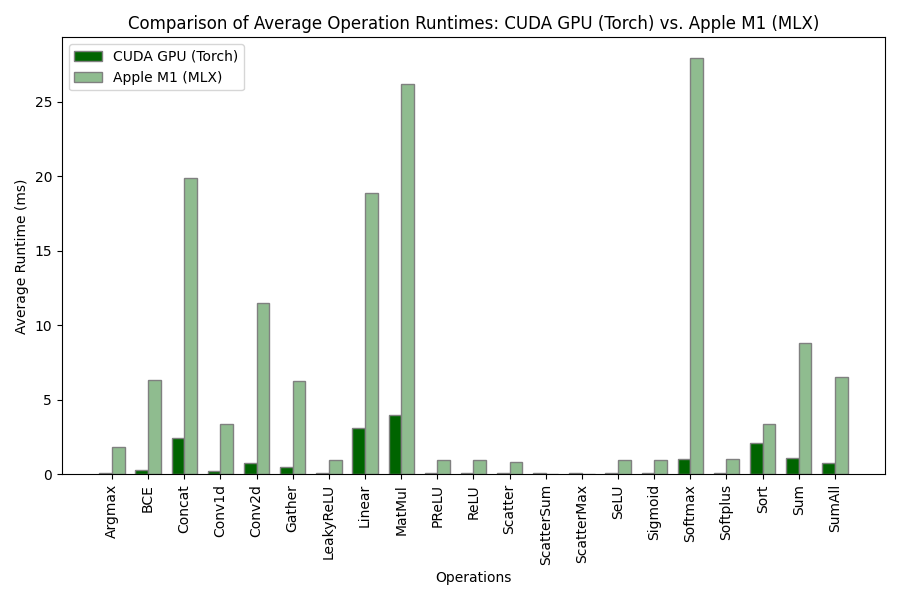}
\caption{Comparison of operation runtimes (in milliseconds) between MLX on Apple M1 GPU and PyTorch on the CUDA GPU. Lower bars indicate better performance. Key transformer operations such as MatMul, Linear, and Softmax are highlighted.}
\label{fig:operation_runtime}
\end{figure}

Our benchmarking of the machine learning operations compared the performance of MLX on Apple Silicon (M1) against PyTorch on the NVIDIA A10 CUDA GPU, focusing on operations crucial for transformer architectures. Figure \ref{fig:operation_runtime} presents a comparison of runtimes for various operations on both devices.

While the CUDA GPU consistently outperformed the Apple M1, the M1's performance remains noteworthy for a consumer device. For instance, matrix multiplication, crucial for attention mechanisms, took 3.96ms on CUDA GPU and 26.19ms on the M1. Linear transformations, essential for projection layers, required 3.11ms on CUDA GPU compared to 18.88ms on the M1. Softmax, used in attention score normalization, took 1.06ms on CUDA GPU and 27.91ms on the M1.

These results demonstrate that while CUDA GPUs maintain a performance edge, Apple Silicon devices offer promising capabilities for on-device machine learning tasks, particularly for research, experimentation, and applications that don't require the absolute highest performance.

\subsection{Benchmarking Model Inference}

Our model inference benchmarking compared MLX performance on Apple Silicon devices (M1 and M2 Max) against PyTorch on a CUDA-enabled GPU (NVIDIA A10). We evaluated BERT (base and large), RoBERTa (base), and XLM-RoBERTa (base) models across various input lengths (50, 100, 200, and 500 characters) and batch sizes (1, 16, and 32).

\begin{figure}[h]
\centering
\includegraphics[width=0.8\textwidth]{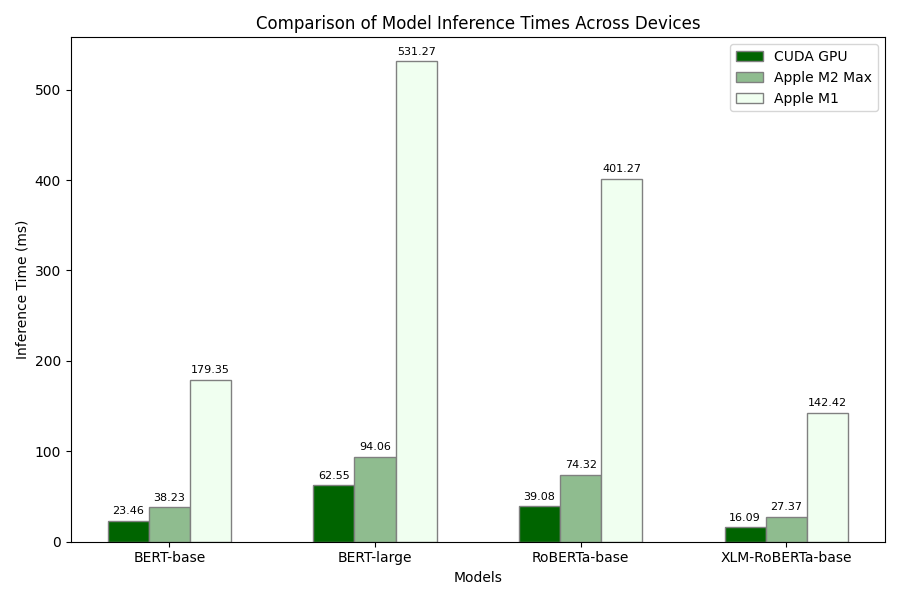}
\caption{Comparison of average inference times (in milliseconds) for different models across CUDA GPU, Apple M1, and Apple M2 Max. Lower bars indicate better performance.}
\label{fig:overall_model_comparison}
\end{figure}

Figure \ref{fig:overall_model_comparison} shows that while the CUDA GPU consistently outperformed both Apple Silicon devices, the M2 Max significantly narrowed the gap compared to the M1. For example, BERT-base inference times were 23.46ms (CUDA), 179.35ms (M1), and 38.23ms (M2 Max). It's worth noting that the M2 Max's superior performance over the M1 is partly due to its better hardware configuration (32GB RAM vs 8GB RAM).

\begin{figure}[h]
\centering
\includegraphics[width=0.8\textwidth]{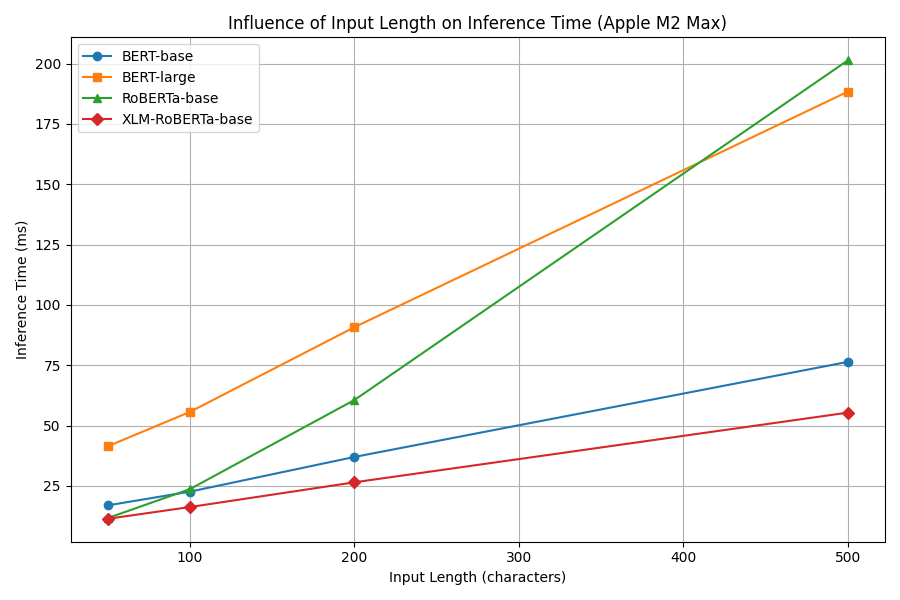}
\caption{Influence of input length on inference times for different models on Apple M2 Max. The x-axis represents input lengths, and the y-axis shows inference times in milliseconds.}
\label{fig:input_length_influence}
\end{figure}

Figure \ref{fig:input_length_influence} illustrates how inference times increased with input length across all models, with varying rates of increase. On the M2 Max, BERT-base showed a relatively linear increase from 16.93ms (50 chars) to 76.40ms (500 chars), while RoBERTa-base exhibited a more pronounced increase from 11.65ms to 201.36ms.

\begin{figure}[h]
\centering
\includegraphics[width=0.8\textwidth]{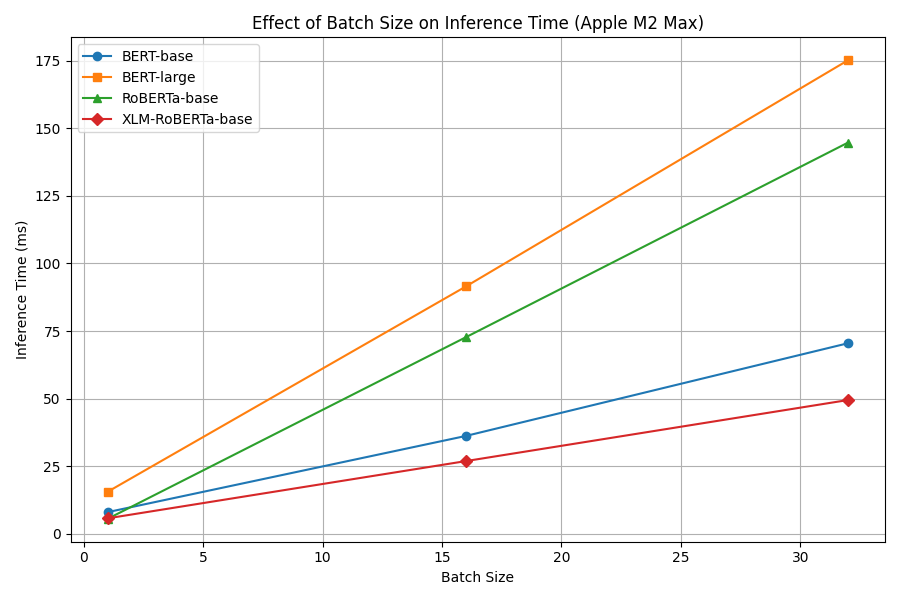}
\caption{Effect of batch size on inference times for different models on Apple M2 Max. The x-axis represents batch sizes, and the y-axis shows inference times in milliseconds.}
\label{fig:batch_size_influence}
\end{figure}

Figure \ref{fig:batch_size_influence} demonstrates the impact of batch size on inference times. All models showed increased inference times with larger batch sizes, but the scaling was not linear. For instance, on the M2 Max, BERT-base inference times increased from 8.02ms (batch size 1) to 70.48ms (batch size 32), an approximately 9x increase for a 32x batch size increase. This sublinear scaling suggests efficient parallel processing capabilities of the M2 Max.

\subsection{Conclusion}

Our benchmarking experiments show that while the CUDA GPU (NVIDIA A10) still lead in overall performance, the Apple Silicon devices we evaluated offer compelling performance for on-device inference.
MLX's optimization for Apple Silicon allows these consumer devices to achieve impressive results, presenting a significant advantage for researchers and developers who may not have access to specialized hardware. 
Notably, we observed a sublinear increase in inference time with batch size on Apple Silicon, indicating good scalability for batch processing tasks. This scalability, combined with consistent performance across various input lengths, positions the MLX-Apple Silicon combination as a viable option for a wide range of machine learning tasks, from research and prototyping to deployment in resource-constrained environments.

The performance of MLX on Apple Silicon suggests promising potential for running large language models on consumer devices. While our current benchmarks focused on models like BERT and RoBERTa, the results indicate we may be approaching a future where more complex models could be run efficiently on personal computers. However, it's important to note that full-scale production workloads remain impractical on consumer devices at present. Nevertheless, the cost-effectiveness of Apple Silicon devices compared to paying for GPU instances on cloud platforms adds to their appeal, allowing researchers and developers to conduct extensive experiments on-device without incurring ongoing cloud computing costs.

In conclusion, while CUDA GPUs remain the preferred choice for performance-critical applications, the combination of MLX and Apple Silicon devices presents a compelling alternative for many machine learning tasks, particularly for on-device inference and experimentation. Their accessibility, cost-effectiveness, and improving performance make them an increasingly attractive choice in the AI and machine learning landscape. This advancement contributes to the democratization of AI technologies, enabling broader access to powerful machine learning capabilities on consumer-grade hardware. Future research should focus on optimizing performance for larger and more diverse models, as well as exploring the potential for on-device training of smaller models, further expanding the capabilities of machine learning on consumer devices.

\clearpage

\bibliographystyle{ACM-Reference-Format}
\balance

\bibliography{main}

\newpage
\appendix

\section*{Appendices} 


{\normalsize These appendices present both the average and detailed runtime benchmarks for various operations and model inferences, measured in milliseconds. The following headers represent different configurations and comparisons used in the analysis:}

\begin{itemize}
{\normalsize \item \textbf{mlx-gpu}: MLX framework using the GPU backend.
\item \textbf{mlx-cpu}: MLX framework using the CPU backend.
\item \textbf{cpu}: PyTorch framework using the CPU backend.
\item \textbf{cuda}: PyTorch framework using the CUDA backend.
\item \textbf{mlx-gpu/mlx-cpu speedup}: The runtime speedup of MLX-GPU compared to MLX-CPU.
\item \textbf{cuda/cpu speedup}: The runtime speedup of CUDA compared to CPU.}
\end{itemize}

\section{Detailed Benchmark Results of the Machine Learning Operations}


\subsection{Benchmark Results: Machine Learning Operations Using PyTorch on Nvidia A10 CUDA GPU}
\insertdetailedoperationbenchmarkcuda
\newpage

\subsection{Benchmark Results: Machine Learning Operations Using MLX on Apple M1 Macbook pro}
\insertdetailedoperationbenchmarkmlx

\newpage
\section{Average Benchmark Results of the Machine Learning Operations}
\subsection{Benchmark Results (Average): Machine Learning Operations Using PyTorch on Nvidia A10 CUDA GPU}
\insertavgoperationbenchmarkcuda
\subsection{Benchmark Results (Average): Machine Learning Operations Using MLX on Apple M1 Macbook pro}
\insertavgoperationbenchmarkmlx
\newpage
\section{Detailed Benchmark Results of the Models Inference Latency}
\subsection{Benchmark Results: Models Inference Latency Using PyTorch on Nvidia A10 CUDA GPU}

\insertdetailedinferencebenchmarkcuda
\newpage

\subsection{Benchmark Results: Models Inference Latency Using MLX on Apple M1 Macbook pro}
\insertdetailedinferencebenchmarkmlx
\newpage
\subsection{Benchmark Results: Models Inference Latency Using MLX on Apple M2 Max Macbook pro}
\insertdetailedinferencebenchmarkmlxmax

\section{Average Benchmark Results of Models Inference Latency based on the Input length}
\subsection{Benchmark Results: Average Models Inference Latency based on the Input length Using PyTorch on Nvidia A10 CUDA GPU}

\insertavginferenceinpbenchmarkcuda
\subsection{Benchmark Results: Average Models Inference Latency based on the Input length Using MLX on Apple M1 Macbook pro}
\insertavginferenceinpbenchmarkmlx
\newpage

\subsection{Benchmark Results: Average Models Inference Latency based on the Input length Using MLX on Apple M2 Max Macbook pro}
\insertavginferenceinpbenchmarkmlxmax
\newpage

\section{Average Benchmark Results of Models Inference Latency based on the Batch Size}
\subsection{Benchmark Results: Average Models Inference Latency based on the Batch Size Using PyTorch on Nvidia A10 CUDA GPU}
\insertavginferencebsizebenchmarkcuda

\subsection{Benchmark Results: Average Models Inference Latency based on the Batch Size Using MLX on Apple M1 Macbook pro}
\insertavginferencebsizebenchmarkmlx
\subsection{Benchmark Results: Average Models Inference Latency based on the Batch Size Using MLX on Apple M2 Max Macbook pro}
\insertavginferencebsizebenchmarkmlxmax
\newpage
\section{Overall Average Benchmark Results of Models Inference Latency}
\subsection{Overall Average Benchmark Results: Models Inference Latency Using PyTorch on Nvidia A10 CUDA GPU}
\insertavgmodelbenchmarkcuda
\subsection{Overall Average Benchmark Results: Models Inference Latency Using MLX on Apple M1 Macbook pro}
\insertavgmodelbenchmarkmlx
\subsection{Overall Average Benchmark Results: Models Inference Latency Using MLX on Apple M2 Max Macbook pro}
\insertavgmodelbenchmarkmlxmax

\end{document}